\documentclass[review,12pt]{elsarticle}

\usepackage{amsmath,amssymb,amsfonts}
\usepackage{graphicx}
\usepackage{float}
\usepackage{bm}
\usepackage{booktabs}
\usepackage{threeparttable}
\usepackage{mathrsfs}
\usepackage{lineno}
\usepackage[hidelinks]{hyperref}

\journal{Knowledge-Based Systems}

\begin{document}
	
	\begin{frontmatter}
		
		\title{GLT-PEFT: Gated Lie-Tucker Parameter-Efficient Fine-Tuning for Alzheimer’s Disease Diagnosis with Hippocampal Segmentation Pretraining}
		
		\author[inst1,inst2]{Guanghua He}		
		
		\author[inst2]{Hancan Zhu\corref{cor1}}
		\ead{hancanzhu@yeah.net}
		
		\author[inst3]{Gaohang Yu\corref{cor2}}
		\ead{maghyu@163.com}
		
		\author[inst1]{An Zhang}
		
		\cortext[cor1]{Corresponding author.}
		\cortext[cor2]{Corresponding author.}
		
		\affiliation[inst1]{
			organization={Department of Mathematics, Hangzhou Dianzi University},
			city={Hangzhou},
			postcode={310018},
			country={China}
		}
		
		\affiliation[inst2]{
			organization={School of Mathematics, Physics and Information, Shaoxing University},
			city={Shaoxing},
			postcode={312000},
			country={China}
		}
		
		\affiliation[inst3]{
			organization={Department of Mathematics, Zhejiang University of Science and Technology},
			city={Hangzhou},
			postcode={310023},
			country={China}
		}

\begin{abstract}
Parameter-efficient fine-tuning (PEFT) has emerged as a promising paradigm for adapting pretrained models under limited data conditions. However, most existing PEFT methods are designed for matrix-structured parameters and are not well suited for high-dimensional convolutional kernels in medical imaging models. Moreover, they typically rely on additive updates and lack mechanisms to preserve the geometric structure of pretrained parameters, while multiplicative (geometry-aware) updates are difficult to integrate within a unified framework.
To address this issue, this paper proposes GLT-PEFT, a gated Lie--Tucker parameter-efficient fine-tuning framework for Alzheimer's disease (AD) diagnosis. The proposed approach transfers a hippocampal segmentation pretrained model to a downstream classification task. Tucker decomposition enables tensor-aware low-rank adaptation of 3D convolutional kernels, while Lie group-based transformations provide structure-preserving multiplicative updates. A gating mechanism further reconciles additive and multiplicative update forms, resulting in a unified and more stable fine-tuning strategy.
Extensive experiments demonstrate that GLT-PEFT achieves effective cross-task transfer while significantly reducing trainable parameters, highlighting its effectiveness for efficient and robust adaptation in medical imaging models.
\end{abstract}


\end{frontmatter}


\section{Introduction}

Deep learning has achieved remarkable success in medical image analysis, particularly in tasks such as segmentation and disease diagnosis \cite{ronneberger2015unet, hatamizadeh2022unetr, khojaste2022ad, liu2020multimodel}. In neuroimaging, accurate characterization of anatomical structures, such as the hippocampus, plays a critical role in understanding and diagnosing Alzheimer's disease (AD), as structural changes in the hippocampus are closely associated with disease progression \cite{halliday2017hippocampal}. Recently, large-scale pretraining on medical image segmentation tasks has emerged as an effective strategy to learn transferable anatomical representations \cite{ma2024medsam}. Such representations provide a strong prior for downstream clinical tasks and are particularly valuable in data-limited scenarios.

However, directly transferring segmentation-pretrained models to downstream tasks, such as AD diagnosis, remains challenging due to task discrepancies and domain shifts \cite{wang2018deep}. While segmentation emphasizes voxel-level structural delineation, diagnosis requires high-level semantic understanding and discriminative feature learning. This mismatch highlights the need for effective adaptation strategies that can preserve anatomical knowledge while facilitating task-specific representation learning. A straightforward solution is to fully fine-tune the pretrained model. Although effective, this approach incurs substantial computational and memory overhead, particularly for 3D medical imaging models with high-dimensional convolutional parameters. Moreover, full fine-tuning is prone to overfitting in small-sample clinical datasets. These limitations have driven the development of parameter-efficient fine-tuning (PEFT) methods, which adapt pretrained models by updating only a small subset of parameters \cite{hu2022lora, houlsby2019adapter}.

Most existing PEFT approaches rely on matrix-structured parameterization and additive updates. While effective for transformer-based models, such designs are less suited to 3D convolutional networks, where parameters naturally exhibit high-order tensor structures. Recent studies have therefore explored tensor-based and geometry-aware adaptation strategies \cite{si2024flora, si2025liera}. Tensor-based methods leverage structured decomposition to capture multi-modal parameter interactions, whereas geometry-aware approaches introduce multiplicative updates that better preserve the underlying parameter manifold. However, these two directions have largely been developed in isolation.

A key challenge arises when attempting to integrate them. Due to the fundamental differences between additive and multiplicative update mechanisms, naive combinations often lead to conflicting optimization dynamics, resulting in unstable training and degraded performance.

To address this issue, we propose a novel parameter-efficient fine-tuning framework, termed GLT-PEFT. The proposed method is built upon a unified Lie–Tucker formulation that jointly incorporates tensor-structured low-rank adaptation and geometry-aware parameter updates. Specifically, Tucker decomposition \cite{tucker1966some, kolda2009tensor} models convolutional parameters in a tensor-aware manner, while Lie group-based transformations \cite{hall2013lie} enable structure-preserving multiplicative updates.

To further resolve the incompatibility between these update forms, we introduce a gated hybrid mechanism that adaptively balances additive and multiplicative (Lie-based) updates. This design provides a unified and stable optimization framework, achieving an improved trade-off between representation capacity and training stability.

We evaluate the proposed GLT-PEFT framework on Alzheimer's disease diagnosis via segmentation-to-diagnosis transfer. Extensive experiments demonstrate that the proposed method improves diagnostic performance while significantly reducing the number of trainable parameters. Moreover, it achieves a favorable sensitivity–specificity balance, which is critical for clinical decision-making.

In summary, the main contributions of this paper are as follows:
\begin{itemize}
	\item We address a fundamental limitation in existing PEFT methods, namely the incompatibility between additive tensor-based updates and multiplicative geometry-aware transformations, which leads to unstable and suboptimal adaptation.
	
	\item We propose GLT-PEFT, a unified Lie--Tucker PEFT framework that integrates tensor-based low-rank adaptation with geometry-preserving multiplicative updates. A gated mechanism is introduced to resolve this inconsistency, enabling stable and effective joint adaptation.
	
	\item We validate the proposed framework on segmentation-to-diagnosis transfer for Alzheimer's disease, showing that GLT-PEFT achieves superior performance with significantly fewer trainable parameters and a favorable sensitivity–specificity balance.
\end{itemize}

\section{Related Work}

\subsection{Parameter-Efficient Fine-Tuning}

Parameter-efficient fine-tuning (PEFT) has emerged as an effective approach for adapting large pretrained models under limited computational resources. A representative method is LoRA \cite{hu2022lora}, which introduces low-rank decomposition to approximate weight updates. Subsequent works improve expressiveness and stability through refined parameterizations. For example, DoRA \cite{liu2024dora} decomposes updates into magnitude and direction components, while PiSSA \cite{meng2024pissa} aligns low-rank subspaces with the principal components of pretrained weights. Orthogonality-constrained methods, such as OPLoRA \cite{oplora2026} and OLoRA \cite{buyukakyuz2024olora}, as well as regularization-based approaches like L2-LoRA \cite{zhang2026l2lora}, further enhance stability and knowledge preservation.

Another line of work focuses on improving generalization and robustness. AdaLoRA \cite{zhang2023adalora} dynamically allocates rank during training, while LoRA-Null \cite{tang2026loranull} mitigates catastrophic forgetting through null-space adaptation. These studies highlight that effective PEFT requires not only efficient parameterization but also careful control of update dynamics.

Beyond low-rank adaptation, recent works explore alternative subspace-based formulations that relax the low-rank assumption. For instance, PMSS \cite{wang2025pmss} selects representative row and column skeletons from pretrained weight matrices and constrains updates within the induced subspace. Compared with low-rank methods, such approaches enable higher-rank updates while preserving intrinsic structural information.

Despite these advances, most existing PEFT methods rely on matrix-based representations and additive updates, which are inherently limited in modeling high-dimensional convolutional kernels. This limitation motivates the development of tensor-structured parameter-efficient adaptation methods.

\subsection{Tensor-Based Parameter-Efficient Adaptation}

To better capture the intrinsic structure of convolutional parameters, tensor-based PEFT methods explicitly model multi-mode weight structures. Classical tensor decompositions, such as Tucker decomposition \cite{tucker1966some,kolda2009tensor} and tensor-train decomposition \cite{oseledets2011tt}, enable structured low-rank representations that preserve correlations across spatial and channel dimensions.

Building on these principles, recent works have extended parameter-efficient adaptation into tensor space using diverse decomposition strategies. Tucker-based approaches, such as FLoRA \cite{si2024flora} and TuckA \cite{lei2026tucka}, construct structured low-rank representations or hierarchical tensor experts to improve adaptation efficiency. Tensor-train-based methods, such as MetaTT \cite{lopez2025metatt}, further model global dependencies across layers through compact tensor-train representations. In addition, LoRA-PT \cite{he2025lorapt} employs tensor singular value decomposition for medical imaging adaptation, while tCURLoRA \cite{he2025tcurlora} leverages tensor CUR decomposition to select informative fibers. TLoRA \cite{tao2025tlora} introduces transform-based tensor parameterization, enhancing flexibility and approximation capability.

Despite these advances, most tensor-based PEFT methods still rely on additive updates in Euclidean parameter space. Such formulations do not explicitly preserve the underlying parameter geometry and may lead to suboptimal optimization behavior, particularly in small-sample medical imaging scenarios. Recent studies on geometry-aware tensor training \cite{zangrando2024geometry} suggest that incorporating manifold structures can improve optimization stability. However, existing tensor-based PEFT methods rarely integrate geometry-aware update mechanisms, indicating that tensor modeling alone is insufficient and motivating the development of unified frameworks that combine tensor-structured representations with geometry-aware adaptations.

\subsection{Geometry-Aware and Lie-Based Fine-Tuning}

Geometry-aware PEFT methods aim to incorporate structural constraints into parameter adaptation by modeling updates on structured manifolds. Early studies primarily introduce orthogonality constraints to restrict updates within complementary subspaces, thereby improving stability and mitigating interference between tasks \cite{wang2023osl}.

More recent advances move beyond conventional Euclidean parameterization and adopt geometry-aware formulations based on manifold structures. In particular, Lie group-based approaches model parameter updates as multiplicative transformations, which preserve the intrinsic structure of model parameters and enable more principled optimization dynamics \cite{si2025liera}. Unlike conventional additive perturbations, such updates ensure that parameters evolve along the underlying manifold, thereby maintaining structural consistency during adaptation. These transformations are typically realized via exponential mappings derived from Lie group theory, which establish a principled connection between the tangent space and the manifold \cite{absil2009optimization}.

Building upon this formulation, subsequent works further introduce additional structural constraints within the Lie group framework. For example, orthogonal low-rank adaptation in Lie groups incorporates both orthogonality and low-rank structures into multiplicative updates, leading to more efficient parameterization while better preserving the intrinsic geometry of model parameters \cite{cao2026oliera}.

Despite these advances, existing geometry-aware methods are primarily designed for matrix-structured parameters and remain largely disconnected from tensor-based representations. In contrast, tensor-based PEFT methods typically rely on additive updates in Euclidean space. This discrepancy between additive and multiplicative update mechanisms leads to distinct optimization behaviors, making their direct integration non-trivial. To address this limitation, we propose a unified framework with a gated hybrid mechanism that adaptively balances additive and multiplicative updates, thereby reconciling their differences and enabling stable and effective parameter-efficient fine-tuning for medical imaging models.

\section{Method}

\subsection{Problem Formulation} \label{sec:problem_formulation}

Let $f_{\theta}$ denote a pretrained segmentation model, where $\theta$ represents parameters learned from a source task (hippocampal segmentation). Let $\mathcal{D} = \{(x_i, y_i)\}_{i=1}^{N}$ denote the downstream dataset for Alzheimer's disease (AD) diagnosis, where $x_i$ is a magnetic resonance imaging (MRI) volume and $y_i \in \{0,1\}$ is the corresponding label.

To enable segmentation-to-diagnosis transfer, we construct a diagnosis model $g_{\theta'}$ by reusing the encoder and bottleneck of $f_{\theta}$ while replacing the task-specific components (e.g., the segmentation decoder) with a lightweight classification head.

The goal is to learn a diagnosis model that predicts
\begin{equation}
	\hat{y} = g_{\theta'}(x),
\end{equation}
where $\theta'$ denotes the adapted model parameters for the downstream task.

Instead of fully fine-tuning $\theta$, we seek a parameter-efficient adaptation strategy that constrains the update within a structured parameter space. Specifically, the adapted parameters are obtained via a structured transformation:
\begin{equation}\label{eq:adapt_function}
	\theta' = \mathcal{T}(\theta; \phi),
\end{equation}
where $\mathcal{T}(\cdot)$ denotes the adaptation function and $\phi$ represents a small set of trainable parameters. In this formulation, $\mathcal{T}$ captures both (i) structural reparameterization of the pretrained model and (ii) parameter-efficient adaptation within a constrained subspace.

The key challenge lies in designing $\mathcal{T}$ such that it satisfies three desirable properties:
\begin{itemize}
	\item \textbf{Structure awareness}: preserving the intrinsic multi-dimensional structure of model parameters;
	\item \textbf{Geometry preservation}: maintaining the underlying relationships of pretrained parameters under structured transformations;
	\item \textbf{Optimization compatibility}: reconciling different update mechanisms to enable stable and effective learning under limited data.
\end{itemize}

To address these challenges, we propose a unified parameter-efficient fine-tuning framework, termed GLT-PEFT, which instantiates $\mathcal{T}$ as a combination of tensor-structured adaptation and geometry-aware transformations, integrated through a gated formulation.

\subsection{Segmentation-to-Diagnosis Transfer Paradigm}

To realize the adaptation function $\mathcal{T}(\theta;\phi)$ in practice, we establish a segmentation-to-diagnosis transfer paradigm based on a pretrained hippocampal segmentation network.

\begin{figure}[t]
	\centering
	\includegraphics[width=\linewidth]{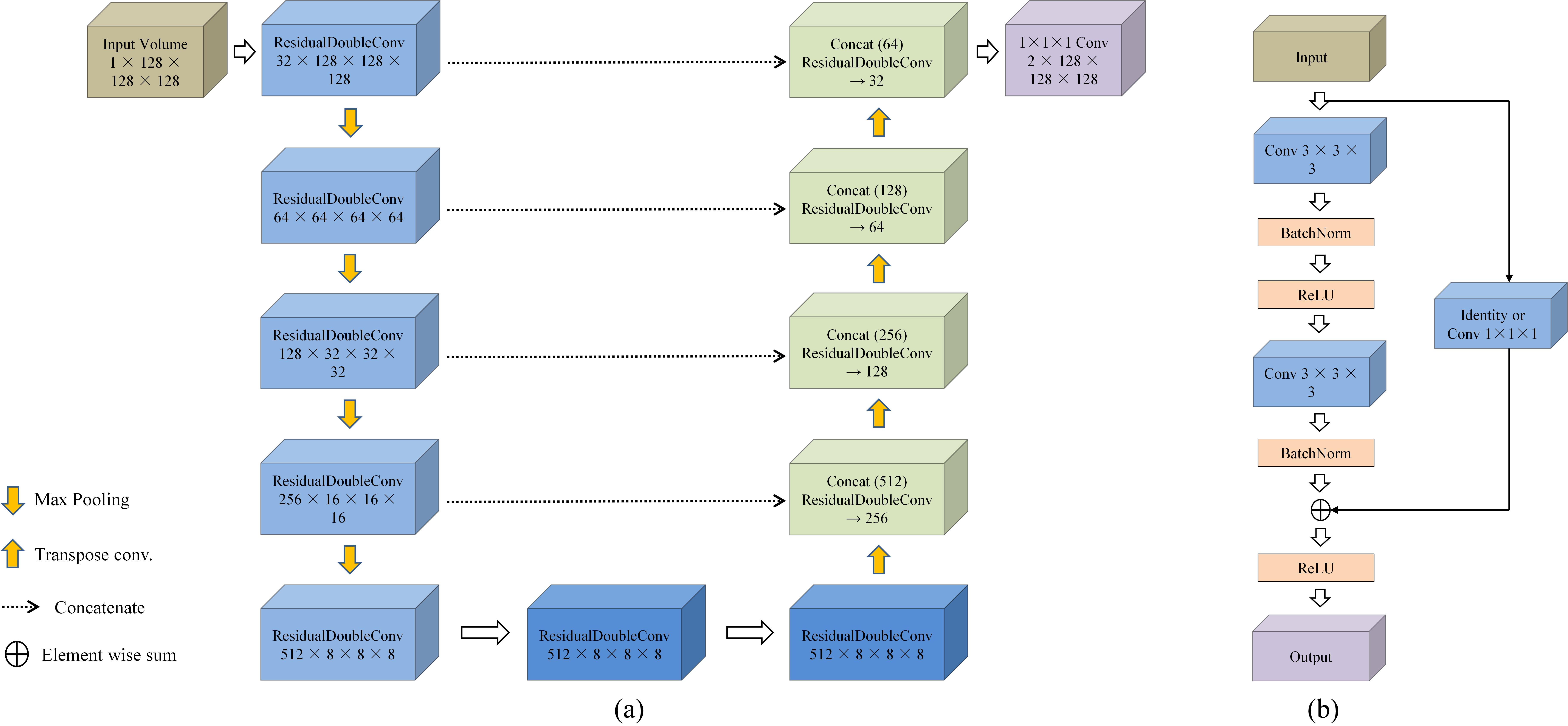}
	\caption{Architecture of the pretrained hippocampal segmentation network. (a) Overall 3D ResUNet-style encoder--decoder framework with an initial stem block, four encoder stages, a bottleneck module, and a symmetric decoder with skip connections. (b) Detailed structure of the ResidualDoubleConv module adopted throughout the network.}
	\label{fig:segmentation}
\end{figure}

The diagnosis model is constructed by reusing the encoder and bottleneck of a pretrained segmentation network (Fig.~\ref{fig:segmentation}a), which follows a 3D ResUNet-style architecture derived from U-Net~\cite{ronneberger2015unet} with residual connections~\cite{he2016resnet}. The backbone consists of an initial stem block, four encoder stages (\texttt{down1--down4}), and a bottleneck module. As illustrated in Fig.~\ref{fig:segmentation}(b), the network is built upon ResidualDoubleConv modules composed of stacked 3D convolutions with residual shortcuts. Since 3D convolution kernels naturally form high-order tensor representations, this architecture provides a natural foundation for tensor-structured parameter-efficient adaptation.

For downstream diagnosis, the task-specific decoder and segmentation head are removed and replaced with a lightweight classification head attached to the bottleneck feature. This transfer paradigm enables the model to reuse anatomically informed representations learned from hippocampal segmentation while adapting to AD diagnosis.

The resulting architecture provides a hierarchical representation space for downstream learning. Specifically, shallow layers capture low-level anatomical structures, whereas deeper layers encode high-level semantic information. Such hierarchical representations are particularly suitable for AD diagnosis, which requires both structural awareness and global contextual reasoning.

Compared with training from scratch, the proposed transfer paradigm improves data efficiency and reduces the risk of overfitting in limited-data medical imaging scenarios. Moreover, it establishes a structured representation foundation for the subsequent tensor-structured and geometry-aware parameter-efficient adaptation.

\begin{figure}[t]
	\centering
	\includegraphics[width=\linewidth]{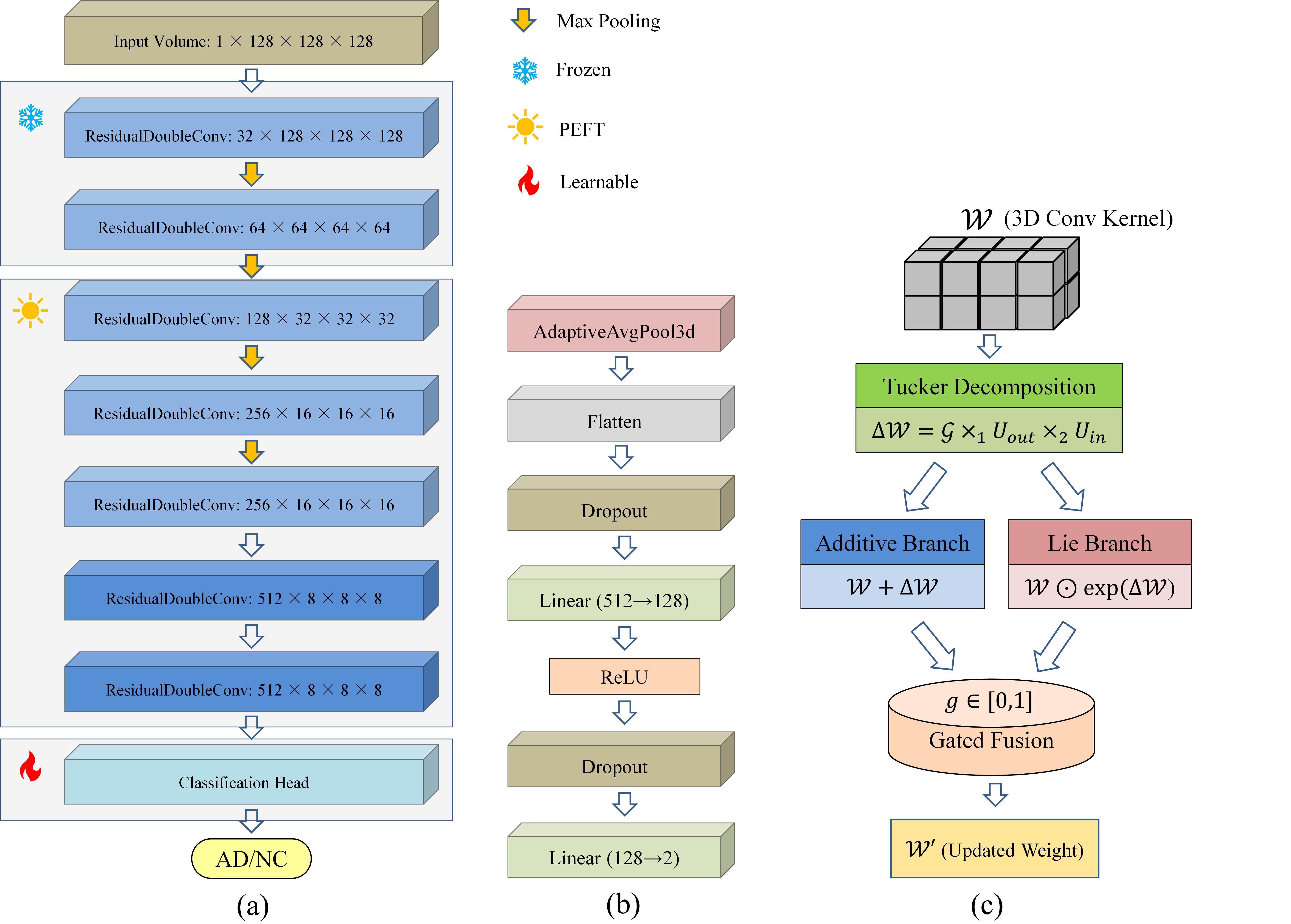}
	\caption{Overview of the proposed GLT-PEFT framework for segmentation-to-diagnosis transfer. (a) Segmentation-to-diagnosis transfer pipeline based on a pretrained hippocampal segmentation backbone. (b) Classification head for downstream Alzheimer's disease (AD) versus normal control (NC) classification. (c) Proposed GLT-PEFT module integrating Tucker-based tensor adaptation, Lie group transformation, and gated fusion.}
	\label{fig:classification}
\end{figure}

\subsection{GLT-PEFT Framework}

To realize the transformation $\mathcal{T}(\theta;\phi)$ within the above transfer paradigm, we propose GLT-PEFT, a unified framework that models parameter updates as a combination of tensor-structured adaptation and geometry-aware transformations.

As illustrated in Fig.~\ref{fig:classification}, GLT-PEFT does not rely on a single update formulation. Instead, it explicitly addresses the fundamental mismatch between additive updates in Euclidean parameter space and multiplicative transformations on parameter manifolds. By jointly modeling the structure of convolutional parameters and the geometry of parameter updates, the proposed framework enables a more flexible and stable adaptation process.

Concretely, GLT-PEFT instantiates $\mathcal{T}(\theta;\phi)$ through three tightly coupled components: (1) Tucker-based low-rank tensor adaptation, (2) Lie group-inspired multiplicative updates, and (3) a hybrid gating mechanism that reconciles their optimization behaviors.

We summarize the proposed update rule as:
\begin{equation}
	\mathscr{W}' = (1-g)(\mathscr{W} + \Delta \mathscr{W}) + g\bigl(\mathscr{W} \odot \exp(\Delta \mathscr{W})\bigr),
\end{equation}
where $\Delta \mathscr{W}$ is obtained from the tensor-structured adaptation, the exponential mapping corresponds to the geometry-aware multiplicative update, and $g \in [0,1]$ is a learnable gate that balances the two update forms.

\subsubsection{Tensor-Structured Adaptation}

To realize the tensor-structured component of the transformation $\mathcal{T}(\theta;\phi)$, we model convolutional parameters using a Tucker-style decomposition that explicitly preserves their multi-dimensional structure.

Let $\mathscr{W} \in \mathbb{R}^{C_{out} \times C_{in} \times k \times k \times k}$ denote a 3D convolutional kernel. Instead of applying matrix-based low-rank adaptation, which flattens parameters and ignores their inherent multi-mode structure, we perform a structured decomposition along the channel dimensions.

Specifically, we factorize only the output and input channel modes while preserving the spatial kernel structure, leading to:
\begin{equation}
	\Delta \mathscr{W} = \mathcal{G} \times_1 U_{out} \times_2 U_{in},
\end{equation}
where $\mathcal{G} \in \mathbb{R}^{r_{out} \times r_{in} \times k \times k \times k}$ is the core tensor, and $U_{out} \in \mathbb{R}^{C_{out} \times r_{out}}$, $U_{in} \in \mathbb{R}^{C_{in} \times r_{in}}$ are factor matrices corresponding to the output and input channel modes, respectively. Here, $\times_n$ denotes the mode-$n$ tensor--matrix product.

This design reflects a key modeling principle: parameter redundancy in convolutional kernels primarily lies in channel interactions, while local spatial patterns should be preserved. Accordingly, the proposed decomposition reduces the number of trainable parameters from $\mathcal{O}(C_{out} C_{in} k^3)$ to $\mathcal{O}(r_{out} r_{in} k^3 + C_{out} r_{out} + C_{in} r_{in})$, while maintaining the spatial expressiveness of the kernel.

Compared with matrix-based PEFT methods, which treat convolutional weights as flattened matrices, the proposed tensor-structured formulation better captures the intrinsic coupling across different modes, leading to more expressive and structurally consistent parameter updates.

\subsubsection{Geometry-Aware Update}

To incorporate geometry-aware transformations into the adaptation function $\mathcal{T}(\theta;\phi)$, we further model parameter updates beyond the conventional additive formulation.

Standard PEFT methods adopt additive updates of the form:
\begin{equation}
	\mathscr{W}' = \mathscr{W} + \Delta \mathscr{W},
\end{equation}
which operate in a linearized Euclidean parameter space. Although simple and effective, such additive perturbations do not explicitly preserve the intrinsic structural relationships of pretrained parameters during adaptation.

To address this limitation, we introduce a Lie group-inspired multiplicative update \cite{si2025liera}:
\begin{equation}
	\mathscr{W}' = \mathscr{W} \odot \exp(\Delta \mathscr{W}),
\end{equation}
where $\odot$ denotes element-wise multiplication. Here, the update $\Delta \mathscr{W}$ is obtained from the tensor-structured adaptation described in the previous subsection.

Unlike additive perturbations, the multiplicative formulation models parameter evolution through structured transformations on a multiplicative manifold. From the perspective of Lie group parameterization, the exponential mapping enables smooth geometry-aware transformations while maintaining relative scaling relationships among pretrained parameters. This property is particularly beneficial for PEFT, where preserving the stability and prior knowledge of pretrained models is critical under limited trainable parameters.

Compared with conventional additive updates, multiplicative updates provide more stable adaptation dynamics and improved robustness, especially in low-data regimes.

For computational efficiency, we adopt the first-order approximation:
\begin{equation}
	\mathscr{W}'  = \mathscr{W} \odot \exp(\Delta \mathscr{W}) \approx \mathscr{W} + \mathscr{W} \odot \Delta \mathscr{W},
\end{equation}
which yields a tractable implementation while retaining the essential characteristics of geometry-aware transformations.

\subsubsection{Gated Unified Update}

Building upon the tensor-structured update $\Delta \mathscr{W}$ and the geometry-aware formulation, we now derive a unified update rule that serves as the final realization of the transformation $\mathcal{T}(\theta;\phi)$.

Although tensor-structured adaptation and geometry-aware updates are individually effective, integrating them within a unified framework is non-trivial due to their fundamentally different update mechanisms. Specifically, tensor-structured adaptation follows an additive formulation in a linearized Euclidean parameter space, whereas the geometry-aware strategy relies on multiplicative transformations that induce non-linear geometric updates. Directly combining these two forms may introduce conflicting optimization dynamics and unstable convergence behavior.

To address this issue, we formulate a unified update rule through a learnable gating mechanism:
\begin{equation}
	\mathscr{W}' = (1-g)(\mathscr{W} + \Delta \mathscr{W}) + g\bigl(\mathscr{W} + \mathscr{W} \odot \Delta \mathscr{W}\bigr),
\end{equation}
where $\Delta \mathscr{W}$ is obtained from the tensor-structured adaptation, and $g \in [0,1]$ is a learnable gate controlling the balance between additive and multiplicative updates.

This formulation provides a unified framework that integrates tensor-structured and geometry-aware adaptation within a single update rule. From an optimization perspective, it can be interpreted as an adaptive interpolation between linear (Euclidean) and manifold-based update regimes.

When $g=0$, the model reduces to purely additive tensor-based adaptation, while $g=1$ corresponds to a fully multiplicative geometry-aware update. Intermediate values enable a smooth transition between the two regimes, allowing the model to dynamically adjust the update strategy during training.

This design provides two key advantages. First, it enables a flexible balance between adaptation flexibility and geometric stability. Second, it alleviates the optimization conflict between additive and multiplicative updates, leading to improved convergence and robustness, particularly in small-sample scenarios.

In practice, we adopt a stage-shared gating strategy, where convolutional layers within the same network stage share a common gate parameter. This design reduces parameter overhead and improves training stability while preserving sufficient adaptability.

\subsection{Optimization and Training Strategy}

The proposed training strategy is designed to optimize the transformation $\mathcal{T}(\theta;\phi)$ while retaining pretrained anatomical knowledge. 
As shown in Fig.~\ref{fig:classification}, we follow a selective adaptation strategy, where shallow layers are frozen to retain stable anatomical features, while GLT-PEFT is applied to deeper convolutional layers to capture task-specific diagnostic patterns. Within the PEFT-adapted layers, lightweight modules, including $1 \times 1 \times 1$ convolutions and normalization layers, are fully trainable to facilitate feature recalibration.

The classification head, denoted as $\Theta_{\text{head}}$, is attached to the bottleneck feature and consists of a global pooling operation followed by a lightweight fully connected classifier.

Formally, the set of trainable parameters can be expressed as
\begin{equation}
	\Theta_{\text{train}} = \Theta_{\text{GLT-PEFT}} \cup \Theta_{\text{light}} \cup \Theta_{\text{head}},
\end{equation}
where $\Theta_{\text{GLT-PEFT}}$ corresponds to the parameters $\phi$ of the transformation $\mathcal{T}(\theta;\phi)$, $\Theta_{\text{light}}$ denotes lightweight trainable modules, and $\Theta_{\text{head}}$ denotes the parameters of the classification head. All remaining parameters are kept fixed.

The downstream objective is to minimize the binary cross-entropy loss:
\begin{equation}
	\mathcal{L}_{cls} = - \frac{1}{N} \sum_{i=1}^{N} \left[ y_i \log \hat{y}_i + (1-y_i)\log(1-\hat{y}_i) \right],
\end{equation}
where $N$ denotes the number of samples, $y_i \in \{0,1\}$ is the ground-truth label of the $i$-th sample, and $\hat{y}_i = \sigma(z_i)$ represents the predicted probability obtained by applying the sigmoid activation function $\sigma(\cdot)$ to the logit $z_i$.

\section{Experiments}

\subsection{Materials}

\subsubsection{Pretraining Dataset}

The segmentation pretraining stage is conducted on the EADC-ADNI dataset with manually annotated hippocampal labels \cite{boccardi2015eadc}, which is derived from the Alzheimer's Disease Neuroimaging Initiative (ADNI) database under a standardized labeling protocol. This dataset provides high-quality anatomical supervision for learning robust hippocampal representations and serves as the foundation for the subsequent segmentation-to-diagnosis transfer.

The dataset contains 135 MRI scans, each with a spatial size of $197 \times 233 \times 189$ and an isotropic voxel spacing of $1 \times 1 \times 1$ mm$^3$. After excluding five scans with annotation inconsistencies, 130 scans were retained for pretraining.

To reduce inter-subject variability and ensure spatial consistency, all scans are processed using a standardized preprocessing pipeline. Specifically, skull stripping is applied to remove non-brain tissues using FreeSurfer \cite{fischl2012freesurfer}, followed by affine registration to the MNI152 standard space using the FLIRT tool from the FSL toolbox \cite{jenkinson2012fsl}. During registration, all scans are resampled to an isotropic resolution of $1 \times 1 \times 1$ mm$^3$ and aligned to a unified spatial size of $182 \times 218 \times 182$. This preprocessing ensures consistent anatomical alignment and facilitates stable training of the 3D segmentation backbone.

\subsubsection{Downstream ADNI Dataset}

The downstream diagnosis task is conducted on a subset of the ADNI database \cite{jack2008adni}. ADNI is a large-scale multi-center longitudinal study designed to investigate the progression of AD through multimodal neuroimaging and clinical assessments. In this work, structural T1-weighted MRI scans from the ADNI2 phase are utilized.

The constructed binary classification dataset consists of 307 subjects, including 145 AD patients and 162 normal controls (NC), where each subject corresponds to one MRI scan. The demographic characteristics of the dataset, including age, gender distribution, education level, and Mini-Mental State Examination (MMSE) scores, are summarized in Table~\ref{tab:adni2_dataset}.

To ensure consistency with the pretraining stage and facilitate effective transfer, the same preprocessing pipeline is applied, including skull stripping, affine registration to the MNI152 standard space, and spatial normalization to a unified resolution and image size.

\begin{table}[t]
	\centering
	\caption{Demographic statistics of the ADNI2 dataset for AD diagnosis.}
	\label{tab:adni2_dataset}
	\small
	\setlength{\tabcolsep}{4pt}
	\begin{threeparttable}
		\begin{tabular}{cccccc}
			\toprule
			Group & Subjects & Female / Male & Age (years) & Education (years) & MMSE \\
			\midrule
			AD & 145 & 59 / 86 & $74.56 \pm 8.17$ & $16.49 \pm 2.66$ & $23.06 \pm 2.09$ \\
			NC & 162 & 91 / 71 & $72.38 \pm 5.78$ & $15.91 \pm 2.52$ & $29.12 \pm 1.18$ \\
			\midrule
			Total & 307 & 150 / 157 & $73.41 \pm 7.08$ & $16.19 \pm 2.61$ & $26.26 \pm 3.46$ \\
			\bottomrule
		\end{tabular}
	\end{threeparttable}
\end{table}

\subsection{Evaluation Metrics}

To comprehensively evaluate the performance of the proposed method for AD diagnosis, we adopt five widely used classification metrics: accuracy (ACC), area under the receiver operating characteristic curve (AUC), sensitivity (SEN), specificity (SPE), and F1-score. In this study, AD patients are treated as the positive class.

Let $TP$, $TN$, $FP$, and $FN$ denote the numbers of true positives, true negatives, false positives, and false negatives, respectively.

Accuracy (ACC) is defined as
\begin{equation}
	\mathrm{ACC} = \frac{TP + TN}{TP + TN + FP + FN}.
\end{equation}

Sensitivity (SEN) and specificity (SPE), which measure the ability to correctly identify positive and negative cases, respectively, are defined as
\begin{equation}
	\mathrm{SEN} = \frac{TP}{TP + FN}, \quad
	\mathrm{SPE} = \frac{TN}{TN + FP}.
\end{equation}

The F1-score is defined as
\begin{equation}
	\mathrm{F1} = \frac{2TP}{2TP + FP + FN}.
\end{equation}

AUC measures the area under the receiver operating characteristic curve and evaluates the discriminative ability of the model across different decision thresholds, thereby providing a threshold-independent assessment of classification performance.

\subsection{Implementation Details}

\paragraph{Pretraining Setup}
The segmentation backbone is pretrained from scratch on the hippocampal segmentation dataset. The dataset is divided into 90 training volumes and 40 validation volumes. The model is trained for 500 epochs with a batch size of 2 using the Adam optimizer. The initial learning rate is set to $1 \times 10^{-4}$ with a weight decay of $2 \times 10^{-5}$. A Dice-based loss function is adopted for hippocampal segmentation. For reproducibility, the random seed is fixed to 1000.

\paragraph{Downstream Task Setup}
Experiments are conducted on the ADNI2 dataset under a binary classification setting (AD vs. NC). Each MRI scan is center-cropped into a fixed-size 3D patch of $128 \times 128 \times 128$. The dataset is divided at the subject level into training, validation, and test sets containing 92, 92, and 123 subjects, respectively, corresponding to an approximate ratio of 3:3:4. This setting is designed to simulate a small-sample scenario with limited training data for downstream adaptation.

During downstream adaptation, the model is trained for 100 epochs using the Adam optimizer with an initial learning rate of $1 \times 10^{-4}$ and a weight decay of $1 \times 10^{-5}$. The batch size is set to 2, and model selection is performed based on validation AUC. All experiments are conducted with a fixed random seed of 1000. All models are implemented in PyTorch and trained on a server equipped with two NVIDIA RTX 4090D GPUs.

\paragraph{Model and PEFT Configuration}

The downstream model is initialized from the pretrained segmentation network by reusing the encoder and bottleneck while replacing the decoder with a lightweight classification head. GLT-PEFT is applied to the $3 \times 3 \times 3$ convolutional layers in the deeper stages (\texttt{down2}, \texttt{down3}, \texttt{down4}, and \texttt{bottleneck}), whereas the shallow stages (\texttt{stem} and \texttt{down1}) remain frozen. For the adapted stages, the corresponding $1 \times 1 \times 1$ shortcut convolutions and affine parameters of normalization layers are fully trainable, together with the classification head. A stage-shared gating strategy is adopted, where convolutional layers within the same stage share a common learnable gate parameter. The gate values are initialized to a constant and optimized during training.

\subsection{Hyperparameter Analysis}

To determine suitable hyperparameter settings for the proposed method, we conduct a grid search over two key hyperparameters: the Tucker rank ratio and the gate initialization value. Model selection is primarily based on validation AUC.

\begin{table}[t]
	\centering
	\caption{Validation AUC of the proposed GLT-PEFT under different rank ratios and gate values.}
	\label{tab:hyperparam_selection}
	\small
	\setlength{\tabcolsep}{4pt}
	\renewcommand{\arraystretch}{1.1}
	\begin{tabular}{c|cccc}
		\toprule
		Rank Ratio $\backslash$ Gate & 0.2 & 0.4 & 0.6 & 0.8 \\
		\midrule
		0.125   & 0.9100 & 0.9176 & 0.9148 & 0.9176 \\
		0.0625  & 0.9110 & 0.9219 & \textbf{0.9280} & 0.9219 \\
		0.03125 & 0.9205 & 0.9205 & 0.9152 & 0.9247 \\
		\bottomrule
	\end{tabular}
\end{table}

As shown in Table~\ref{tab:hyperparam_selection}, the best performance is achieved with a rank ratio of 0.0625 and a gate value of 0.6, yielding the highest validation AUC of 0.9280. This result suggests that an intermediate rank ratio provides a favorable balance between adaptation flexibility and parameter efficiency, while a moderate gate value effectively reconciles additive and multiplicative update behaviors. Based on this configuration, we adopt $\mathrm{ratio}=0.0625$ and $g=0.6$ in subsequent experiments.

In addition to the optimal configuration, we additionally consider an efficiency-oriented setting (ratio = 0.03125, gate = 0.8), which achieves the second-best validation AUC (0.9247) with a smaller rank ratio. This configuration is used for further efficiency analysis.

For fair comparison, hyperparameters of baseline methods are selected using the same validation protocol. For matrix-based PEFT methods (LoRA \cite{hu2022lora}, PiSSA \cite{meng2024pissa}, PMSS \cite{wang2025pmss}, and LieRA \cite{si2025liera}), the rank parameter $r$ is selected from $\{2, 4, 8, 16, 32\}$. Model selection is primarily based on validation AUC, while SEN and SPE are additionally examined to avoid configurations with severely imbalanced diagnostic performance. For the tensor-based baseline FLoRA \cite{si2024flora}, the rank ratio is selected from $\{0.125, 0.0625, 0.03125\}$ using the same validation protocol.

The final hyperparameter settings are as follows: LoRA uses $r=32$, PiSSA, PMSS, and LieRA use $r=8$, and FLoRA uses $\mathrm{ratio}=0.125$. These settings are used in all subsequent comparison experiments.

\subsection{Main Results}

\begin{table}[t]
	\centering
	\caption{Comparison of classification performance on the ADNI test set. Boldface indicates the best performance in each metric.}
	\label{tab:main_results}
	\small
	\setlength{\tabcolsep}{4pt}
	\renewcommand{\arraystretch}{1.1}
	\begin{tabular}{lccccc}
		\toprule
		Method & AUC & ACC & F1 & SEN & SPE \\
		\midrule
		FT & 0.8912 & 0.8130 & 0.7965 & \textbf{0.7759} & 0.8462 \\
		LoRA \cite{hu2022lora} & 0.9167 & 0.8374 & 0.7917 & 0.6552 & \textbf{1.0000} \\
		PiSSA \cite{meng2024pissa} & 0.9210 & 0.8537 & 0.8235 & 0.7241 & 0.9692 \\
		PMSS \cite{wang2025pmss} & 0.9088 & 0.7073 & 0.5500 & 0.3793 & \textbf{1.0000} \\
		LieRA \cite{si2025liera} & 0.8947 & 0.7967 & 0.7573 & 0.6724 & 0.9077 \\
		FLoRA \cite{si2024flora} & 0.9212 & 0.8537 & 0.8235 & 0.7241 & 0.9692 \\
		GLT-PEFT & \textbf{0.9427} & \textbf{0.8862} & \textbf{0.8654} & \textbf{0.7759} & 0.9846 \\
		\bottomrule
	\end{tabular}
\end{table}

Table~\ref{tab:main_results} compares the proposed GLT-PEFT with full fine-tuning and several representative PEFT methods on the AD diagnosis task. Both overall and clinically relevant metrics are included for comprehensive evaluation.

\textbf{(1) Comparison with FT and PEFT baselines.} GLT-PEFT achieves the best  performance across most evaluation metrics, with the highest AUC (0.9427), ACC (0.8862), and F1-score (0.8654), while matching the best SEN (0.7759). Compared with FT, GLT-PEFT improves all evaluation metrics. In particular, SPE is substantially improved (0.8462 $\rightarrow$ 0.9846) while maintaining the same SEN, indicating enhanced generalization ability without sacrificing sensitivity to positive cases.

Among the PEFT baselines, PiSSA and FLoRA achieve relatively strong and stable performance. LoRA attains high AUC and perfect SPE but exhibits lower SEN (0.6552), suggesting reduced sensitivity to AD subjects. PMSS shows the weakest overall performance, particularly in SEN (0.3793), indicating overly conservative predictions. LieRA improves upon PMSS but remains inferior to the stronger PEFT baselines.

\textbf{(2) SEN--SPE trade-off.} GLT-PEFT maintains a more balanced trade-off between SEN and SPE compared with existing PEFT methods. Unlike LoRA and PMSS, which favor specificity at the expense of sensitivity, the proposed method achieves high specificity while preserving strong sensitivity to AD cases.

Overall, these results indicate that the proposed integration of tensor-structured adaptation and geometry-aware updates provides a more effective and balanced parameter-efficient adaptation strategy for AD diagnosis.

\subsection{Computational Efficiency}

\begin{table}[t]
	\centering
	\caption{Comparison of performance and computational efficiency on the ADNI test set, including an efficiency-oriented variant of GLT-PEFT.}
	\label{tab:efficiency_combined}
	\setlength{\tabcolsep}{4pt}
	\renewcommand{\arraystretch}{1.1}
	\begin{threeparttable}
		\resizebox{\columnwidth}{!}{%
			\begin{tabular}{lccccc}
				\toprule
				Method & AUC & ACC & Params & Time (ms) & Memory (MB) \\
				\midrule
				FT & 0.8912 & 0.8130 & 42.69M & 61.96 & 2602 \\
				LoRA \cite{hu2022lora} & 0.9167 & 0.8374 & 3.50M & 59.12 & 2491 \\
				PiSSA \cite{meng2024pissa} & 0.9210 & 0.8537 & 1.21M & 59.04 & 2472 \\
				PMSS \cite{wang2025pmss} & 0.9088 & 0.7073 & 0.45M & 58.92 & 2472 \\
				LieRA \cite{si2025liera} & 0.8947 & 0.7967 & 1.21M & 59.68 & 2467 \\
				FLoRA \cite{si2024flora} & 0.9212 & 0.8537 & 1.51M & 184.73 & 17917 \\
				\midrule
				\textbf{GLT-PEFT} & 0.9427 & \textbf{0.8862} & 0.61M & 62.18 & 4723 \\
				\textbf{GLT-PEFT (efficient)} & \textbf{0.9446} & 0.8618 & \textbf{0.39M} & \textbf{39.35} & \textbf{1974} \\
				\bottomrule
			\end{tabular}%
		}
	\end{threeparttable}
\end{table}

Table~\ref{tab:efficiency_combined} compares the proposed GLT-PEFT with FT and representative PEFT methods in terms of predictive performance, trainable parameters, training time, and peak training memory. 
Although most PEFT methods substantially reduce the number of trainable parameters compared with FT, the corresponding reductions in training time and memory consumption are relatively limited. This observation suggests that parameter reduction alone does not necessarily translate into practical computational efficiency.

Under the default configuration selected by validation AUC, GLT-PEFT achieves strong predictive performance with moderate computational cost. In particular, the proposed method requires only 0.61M trainable parameters, which is substantially lower than FT (42.69M) and is also lower than that of most PEFT baselines.

To further investigate the efficiency--performance trade-off, we additionally evaluate an efficiency-oriented configuration of GLT-PEFT (ratio = 0.03125, gate = 0.8). This setting reduces training time from 62.18 ms to 39.35 ms and peak memory consumption from 4723 MB to 1974 MB, while maintaining strong predictive performance (AUC = 0.9446, ACC = 0.8618). For clarity, this configuration is denoted as GLT-PEFT (efficient) in Table~\ref{tab:efficiency_combined}. Notably, the efficient configuration achieves slightly higher AUC than the default setting, suggesting that a more constrained parameterization may provide additional regularization benefits under limited training data.

Compared with matrix-based PEFT methods such as LoRA, PiSSA, and LieRA, which mainly reduce parameter counts while providing relatively limited reductions in training time and memory consumption, GLT-PEFT (efficient) achieves a more favorable balance between predictive performance and computational cost.
Moreover, unlike the tensor-based baseline FLoRA, which introduces substantial training overhead in both memory and computation, {GLT-PEFT (efficient) maintains high predictive performance while significantly improving practical training efficiency.

\subsection{Ablation Studies}

To systematically evaluate the proposed method, we conduct ablation studies from four perspectives: trainable components, adaptation depth, gating strategy, and model components.

\begin{table}[t]
	\centering
	\caption{Ablation study of trainable components on the ADNI test set.}
	\label{tab:effect_ablation_updated}
	\small
	\setlength{\tabcolsep}{4pt}
	\renewcommand{\arraystretch}{1.1}
	\begin{tabular}{lccccc}
		\toprule
		Configuration & AUC & ACC & F1 & SEN & SPE \\
		\midrule
		Head Only & 0.6764 & 0.5285 & 0.0000 & 0.0000 & 1.0000 \\
		Conv1$\times$1 + Norm + Head & 0.8849 & 0.8211 & 0.8036 & \textbf{0.7759} & 0.8615 \\
		PEFT + Head & 0.9225 & 0.8293 & 0.7921 & 0.6897 & 0.9538 \\
		GLT-PEFT & \textbf{0.9427} & \textbf{0.8862} & \textbf{0.8654} & \textbf{0.7759} & \textbf{0.9846} \\
		\bottomrule
	\end{tabular}
\end{table}

\subsubsection{Ablation of Trainable Components}

We first evaluate the contribution of different trainable components on the ADNI classification task. In the complete GLT-PEFT configuration, GLT-PEFT is applied to the $3 \times 3 \times 3$ convolutional layers in \texttt{down2--down4} and the bottleneck, while the corresponding $1 \times 1 \times 1$ convolutional layers, normalization affine parameters in these stages, and the classification head remain fully trainable. Based on different combinations of trainable components, we construct four variants: \emph{Head Only}, \emph{Conv1$\times$1 + Norm + Head}, \emph{PEFT + Head}, and \emph{GLT-PEFT}, where \emph{GLT-PEFT} denotes the complete configuration. The results are summarized in Table~\ref{tab:effect_ablation_updated}.

As shown in Table~\ref{tab:effect_ablation_updated}, the \emph{Head Only} variant performs poorly, with an AUC of 0.6764 and zero sensitivity, indicating failure to identify positive cases. This result suggests that adapting only the classification head is insufficient for effective segmentation-to-diagnosis transfer.

Introducing lightweight trainable components (\emph{Conv1$\times$1 + Norm + Head}) leads to substantial performance improvement, demonstrating the importance of feature adaptation beyond the classification head alone. The \emph{PEFT + Head} variant achieves a notably higher AUC (0.9225), indicating the effectiveness of structured parameter adaptation for downstream diagnosis.

The \emph{GLT-PEFT} configuration achieves the best performance across all evaluation metrics, demonstrating that jointly adapting structured PEFT modules and lightweight trainable components leads to more effective downstream adaptation.

\begin{table}[t]
	\centering
	\caption{Ablation of adaptation depth on the ADNI test set.}
	\label{tab:structure_ablation_updated}
	\small
	\setlength{\tabcolsep}{4pt}
	\renewcommand{\arraystretch}{1.1}
	\begin{tabular}{lccccc}
		\toprule
		Configuration & AUC & ACC & F1 & SEN & SPE \\
		\midrule
		Bottleneck Only & 0.8997 & 0.7480 & 0.6437 & 0.4828 & \textbf{0.9846} \\
		Default (mid--deep) & \textbf{0.9427} & \textbf{0.8862} & \textbf{0.8654} & \textbf{0.7759} & \textbf{0.9846} \\
		All Stages & 0.9321 & 0.8699 & 0.8491 & \textbf{0.7759} & 0.9538 \\
		\bottomrule
	\end{tabular}
\end{table}

\subsubsection{Ablation on Adaptation Depth}

We next investigate the impact of applying GLT-PEFT at different adaptation depths. Three configurations are considered: \emph{Bottleneck Only}, \emph{Default (mid--deep)}, and \emph{All Stages}, where “mid--deep” refers to applying GLT-PEFT to \texttt{down2--down4} and the bottleneck. The results are summarized in Table~\ref{tab:structure_ablation_updated}.

As shown in Table~\ref{tab:structure_ablation_updated}, the \emph{Default (mid--deep)} configuration achieves the best overall performance, attaining the highest AUC, ACC, and F1 while maintaining strong SEN and SPE. This result indicates that adapting middle-to-deep encoder stages provides an effective balance between task-specific adaptation and representation stability.

In contrast, the \emph{Bottleneck Only} variant exhibits a substantially lower SEN (0.4828), suggesting that adapting only the deepest features is insufficient to capture discriminative diagnostic information. Extending PEFT to \emph{All Stages} slightly degrades performance, indicating that excessive adaptation of shallow layers may disrupt stable low-level representations.

These results suggest that restricting adaptation to middle-to-deep layers is more effective than adapting only the bottleneck or the entire encoder.

\begin{table}[t]
	\centering
	\caption{Ablation of gate mechanisms on the ADNI test set.}
	\label{tab:gate_ablation_updated}
	\small
	\setlength{\tabcolsep}{4pt}
	\renewcommand{\arraystretch}{1.1}
	\begin{tabular}{lccccc}
		\toprule
		Gate Mechanism & AUC & ACC & F1 & SEN & SPE \\
		\midrule
		Fixed Gate & 0.9353 & 0.8780 & 0.8571 & \textbf{0.7759} & 0.9692 \\
		Learnable Gate & 0.9305 & 0.8699 & 0.8491 & \textbf{0.7759} & 0.9538 \\
		Stage-Shared Learnable Gate & \textbf{0.9427} & \textbf{0.8862} & \textbf{0.8654} & \textbf{0.7759} & \textbf{0.9846} \\
		\bottomrule
	\end{tabular}
\end{table}

\subsubsection{Ablation of Gate Mechanism}

We further investigate the effect of different gating strategies in GLT-PEFT. Three variants are considered: \emph{Fixed Gate}, \emph{Learnable Gate}, and \emph{Stage-Shared Learnable Gate}. The fixed gate remains constant during training, the learnable gate assigns an independent parameter to each layer, and the stage-shared strategy shares a single gate parameter within each stage to improve optimization stability.

Under a fixed hyperparameter configuration (rank ratio = 0.0625, initial gate = 0.6), we compare these variants. The results are summarized in Table~\ref{tab:gate_ablation_updated}.
The \emph{Stage-Shared Learnable Gate} achieves the best overall performance. The \emph{Fixed Gate} slightly outperforms the fully learnable variant, suggesting that excessive layer-wise flexibility may introduce unstable optimization dynamics in small-sample settings. In contrast, the stage-shared strategy provides a more structured parameterization, enabling adaptive balancing between additive and multiplicative updates while maintaining stable training behavior. This suggests that moderate parameter sharing is beneficial for stable and effective gate optimization.

\begin{table}[t]
	\centering
	\caption{Component-wise ablation on the ADNI test set.}
	\label{tab:ablation_components}
	\small
	\setlength{\tabcolsep}{5pt}
	\renewcommand{\arraystretch}{1.1}
	\begin{tabular}{lccccc}
		\toprule
		Method & Tucker & Lie & Gate & AUC & ACC \\
		\midrule
		LieRA &  & $\checkmark$ &  & 0.8947 & 0.7967 \\
		FLoRA (Tucker) & $\checkmark$ &  &  & 0.9212 & 0.8537 \\
		Tucker + Lie & $\checkmark$ & $\checkmark$ &  & 0.8920 & 0.8130 \\
		GLT-PEFT & $\checkmark$ & $\checkmark$ & $\checkmark$ & \textbf{0.9427} & \textbf{0.8862} \\
		\bottomrule
	\end{tabular}
\end{table}

\subsubsection{Component-wise Ablation}

To better understand the contribution of each component in the proposed method, we conduct a component-wise ablation study focusing on Tucker-based tensor modeling, Lie group transformation, and the gating mechanism. Specifically, we compare four variants: FLoRA based on Tucker tensor decomposition, LieRA based on Lie group transformation, a direct combination of Tucker modeling and Lie transformation (\emph{Tucker + Lie}), and the GLT-PEFT framework with a stage-shared learnable gating mechanism. For clarity, we report AUC and ACC. The results are summarized in Table~\ref{tab:ablation_components}.

As shown in Table~\ref{tab:ablation_components}, the Tucker-based variant FLoRA achieves strong performance (AUC = 0.9212), substantially outperforming the Lie-based baseline LieRA (AUC = 0.8947). This result indicates that modeling convolutional weights in a tensor-structured manner is effective for downstream adaptation.

However, directly combining Tucker modeling with Lie-based transformation (\emph{Tucker + Lie}) leads to degraded performance (AUC = 0.8920, ACC = 0.8130), suggesting that additive tensor adaptation and multiplicative geometry-aware updates may exhibit incompatible optimization behaviors when naively integrated.

After introducing the stage-shared learnable gating mechanism, GLT-PEFT achieves the best overall performance (AUC = 0.9427, ACC = 0.8862). This result suggests that the gating mechanism effectively reconciles the interaction between Tucker-based adaptation and Lie-based transformation, enabling more stable and effective optimization.

\section{Discussion}
\subsection{Optimization Behavior Analysis}

We analyze the magnitude of parameter updates introduced by different fine-tuning strategies to gain insight into the optimization behavior of the proposed method. Specifically, for each selected convolutional layer (\textit{down2}--\textit{down4} and \textit{bottleneck}), we compute the relative update magnitude using the Frobenius norm,
\[
\|\Delta W\|_F / \|W\|_F,
\]
where $\Delta W$ denotes the effective update with respect to the pretrained weights. The reported values correspond to the average relative update magnitude across all selected layers.

Figure~\ref{fig:deltaW_comparison} compares the update magnitudes across different methods. FT produces the largest relative updates, indicating a highly flexible but weakly constrained optimization process. Such large updates may increase the risk of overfitting, particularly in limited-data settings.

In contrast, conventional PEFT methods, such as LoRA and PiSSA, yield substantially smaller updates. While this improves stability and parameter efficiency, it may also restrict the model's capacity to adapt to downstream tasks. Methods such as PMSS and LieRA further exhibit extremely small updates, suggesting that overly constrained parameterization may limit adaptation capability.

Notably, GLT-PEFT exhibits a moderate update magnitude, positioned between FT and existing PEFT methods. This indicates that it achieves a balance between adaptation flexibility and optimization stability, enabling effective adaptation without excessive parameter changes. Compared with the Tucker-based baseline FLoRA, which produces relatively larger updates, GLT-PEFT achieves more controlled update magnitudes through the proposed gating mechanism. This highlights the role of the interaction between Lie-based transformation and gating in regulating update dynamics.

\begin{figure}[t]
	\centering
	\includegraphics[width=0.8\linewidth]{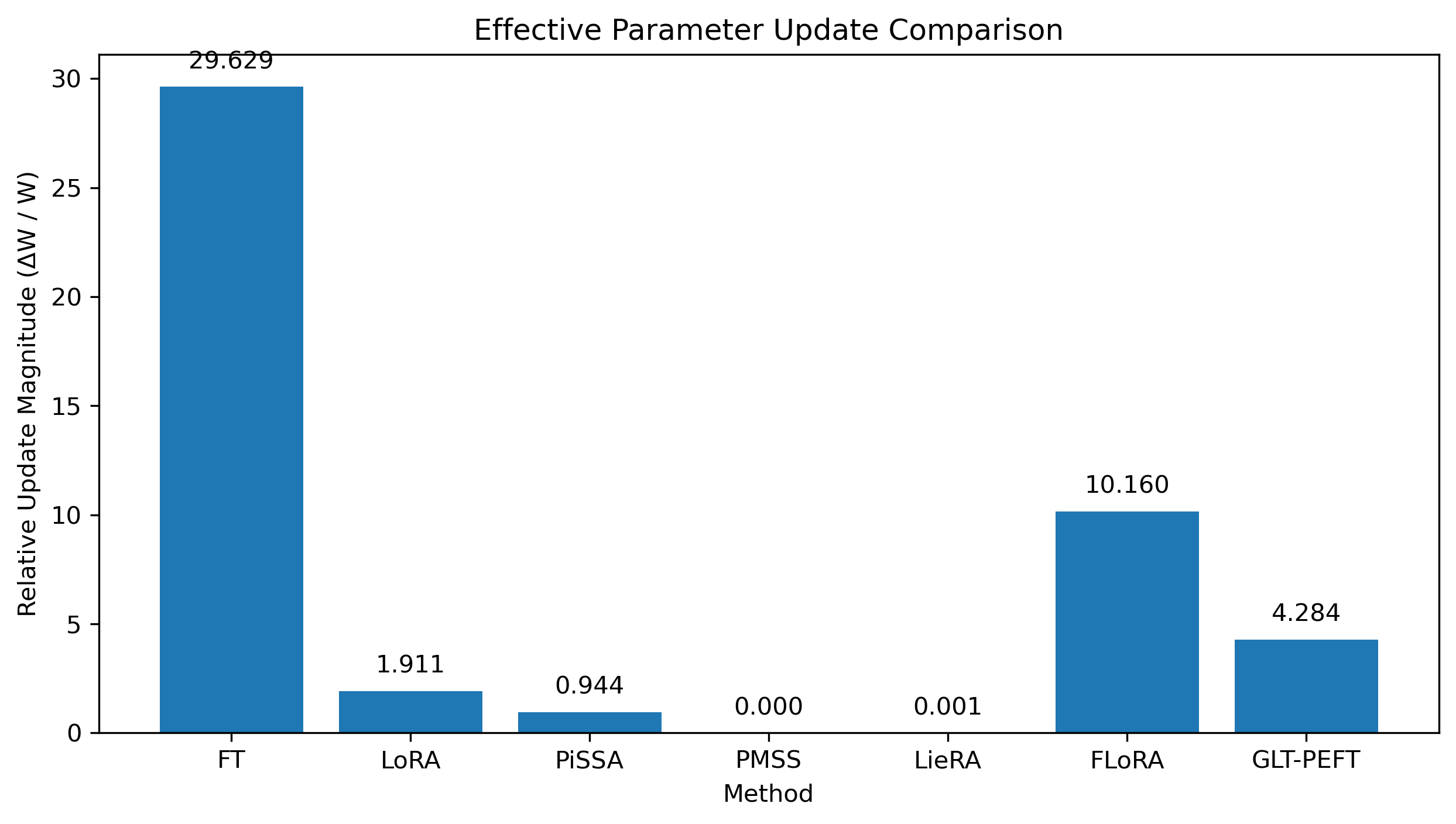}
\caption{Comparison of relative parameter update magnitudes 
	($\|\Delta W\|_F / \|W\|_F$) across different fine-tuning methods. 
	The reported values correspond to the average relative update magnitude across selected convolutional layers.}
	\label{fig:deltaW_comparison}
\end{figure}

\begin{figure}[t]
	\centering
	\includegraphics[width=\linewidth]{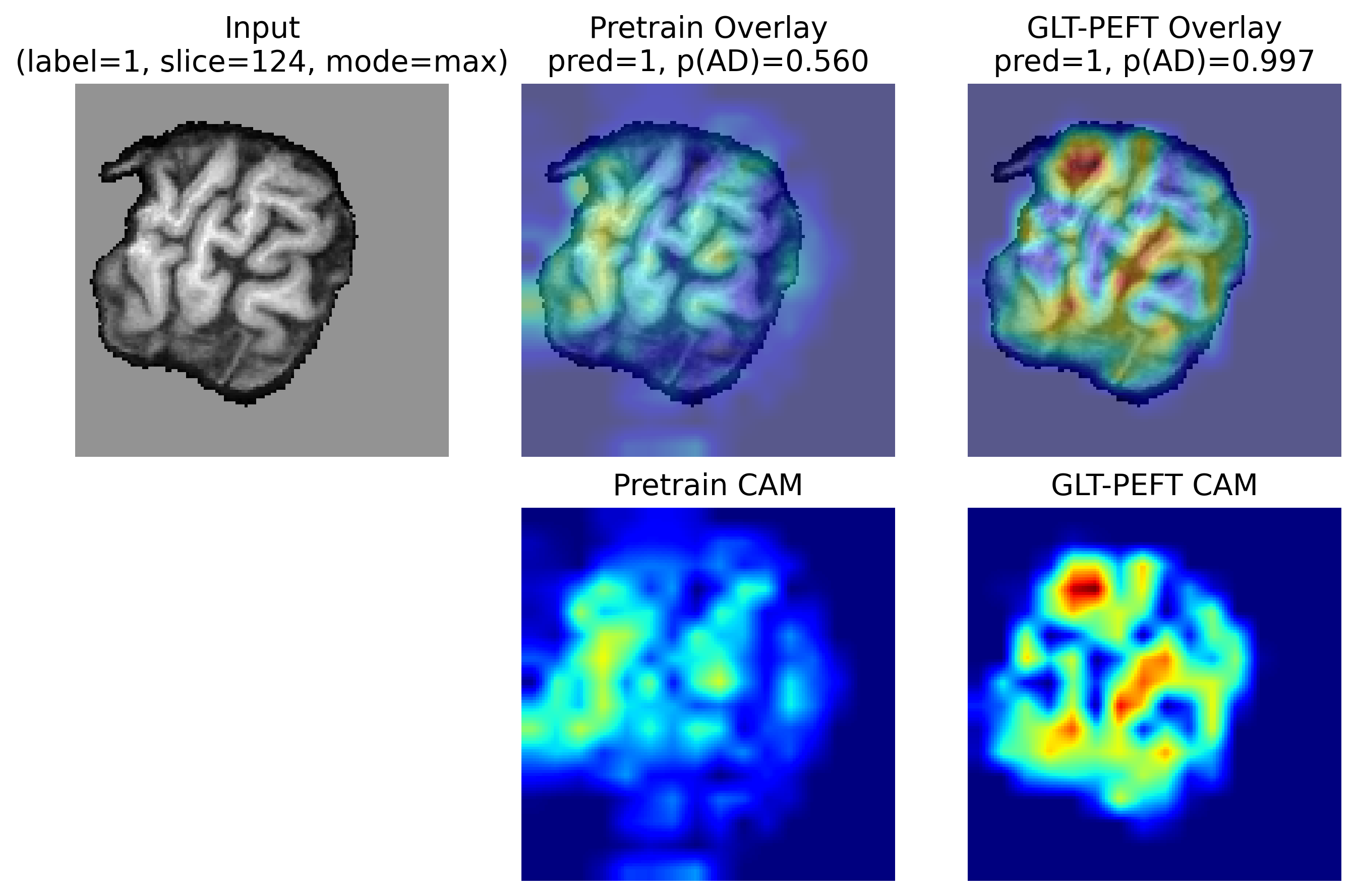}
	\caption{Grad-CAM visualization of a representative AD sample. Compared with the pretrained backbone, GLT-PEFT produces more localized and spatially coherent activation patterns.}
	\label{fig:gradcam_comparison}
\end{figure}

\subsection{Analysis of Feature Representations via Visualization}

To further analyze the learned feature representations, we visualize model responses using Grad-CAM \cite{selvaraju2017gradcam}. As shown in Fig.~\ref{fig:gradcam_comparison}, the pretrained backbone exhibits diffuse and spatially less focused activation patterns, with responses distributed across broad regions. This suggests limited adaptation to the downstream AD diagnosis task.

In contrast, GLT-PEFT produces more concentrated and spatially coherent activation regions. The responses are more localized and exhibit clearer structural patterns, while activations in less relevant regions are suppressed. This indicates improved ability to capture more discriminative diagnostic patterns.

This behavior is consistent with the effect of structured parameter-efficient adaptation. Tucker-based modeling constrains updates within a low-rank subspace, reducing redundant variations, while Lie-based transformation introduces structured geometric updates that enrich adaptation patterns. The gating mechanism further balances these components, leading to more stable and discriminative feature learning.

The prediction confidence is also substantially improved (from 0.560 to 0.997), suggesting that the refined feature representations support more confident predictions. Similar qualitative trends are observed across additional samples.

\subsection{Limitations and Future Work}

Despite the promising results, several limitations remain. The generalization ability of the proposed method under heterogeneous and clinically complex scenarios has yet to be fully validated. Extending the evaluation to more challenging settings, such as multi-class diagnosis, cross-domain variations, and multi-center data, remains an important direction for future study.

On the methodological side, although GLT-PEFT integrates Tucker-based tensor adaptation and Lie group transformations, its theoretical properties are not yet fully understood. In particular, the optimization dynamics within structured low-rank subspaces and the interaction between additive and multiplicative updates under the gating mechanism require further investigation.

The Tucker rank ratio and gate parameter are currently selected through grid search based on validation performance. While effective, this strategy does not adapt dynamically to different data conditions. Data-driven approaches, such as adaptive rank learning or dynamic gate scheduling, may further improve robustness and efficiency.

Future work will focus on strengthening the theoretical foundation of tensor-structured parameter-efficient fine-tuning, particularly from the perspectives of subspace optimization and optimization stability, while further improving robustness under distribution shifts and enabling adaptive parameter selection.

\section{Conclusion}

This paper presents GLT-PEFT, a parameter-efficient fine-tuning framework designed to reconcile additive tensor-based adaptation and multiplicative geometry-aware transformations. By integrating Tucker-based tensor decomposition and Lie group transformations within a unified gated formulation, the proposed method enables structured and stable parameter adaptation under limited-data conditions.
Extensive experiments on Alzheimer's disease diagnosis demonstrate that GLT-PEFT consistently outperforms existing PEFT methods. The proposed approach achieves superior predictive performance while substantially reducing the number of trainable parameters and maintaining a favorable sensitivity--specificity balance. Ablation studies and feature visualization further show that the interaction between tensor modeling, Lie-based transformation, and the stage-shared learnable gating mechanism is critical for effective downstream adaptation.
These findings highlight the potential of combining tensor-structured adaptation with geometry-aware optimization for parameter-efficient fine-tuning in medical image analysis.

\bibliographystyle{elsarticle-num}
\bibliography{peft_bibtex}
\end{document}